\pdfoutput=1

\documentclass[11pt]{article}

\usepackage{emnlp2021}

\usepackage{times}
\usepackage{latexsym}

\usepackage[T1]{fontenc}

\usepackage[utf8]{inputenc}

\usepackage{microtype}

\usepackage{amsmath}
\usepackage{booktabs}
\usepackage{enumitem}
\usepackage{graphicx}
\usepackage{tabularx}
\usepackage{todonotes}
\usepackage{xspace}

\newcommand{\distillingeval}{\mbox{DistilLingEval}\xspace}
\newcommand{\lingeval}{\mbox{LingEval97}\xspace}

\title{On the Limits of Minimal Pairs in Contrastive Evaluation}

\author{Jannis Vamvas$^1$ \and Rico Sennrich$^{1,2}$\\
  $^1$Department of Computational Linguistics, University of Zurich\\
  $^2$School of Informatics, University of Edinburgh \\ \medskip
  \texttt{\{vamvas,sennrich\}@cl.uzh.ch}}

\begin{document}
\maketitle
\begin{abstract}
Minimal sentence pairs are frequently used to analyze the behavior of language models.
It is often assumed that model behavior on contrastive pairs is predictive of model behavior at large.
We argue that two conditions are necessary for this assumption to hold:
First, a tested hypothesis should be well-motivated, since experiments show that contrastive evaluation can lead to false positives.
Secondly, test data should be chosen such as to minimize distributional discrepancy between evaluation time and deployment time.
For a good approximation of deployment-time decoding, we recommend that minimal pairs are created based on machine-generated text, as opposed to human-written references.
We present a contrastive evaluation suite for English--German MT that implements this recommendation.\footnote{\url{https://github.com/ZurichNLP/distil-lingeval}}
\end{abstract}

\section{Introduction}

Contrastive evaluation is one of the most widely used evaluation techniques for generative language models, both for causal models~\cite{linzen-etal-2016-assessing} and sequence-to-sequence models~\cite{sennrich-2017-grammatical}.
Various phenomena have been analyzed using this technique, including
syntax (\citealt{marvin-linzen-2018-targeted}; among others),
word sense disambiguation (\citealt{rios-gonzales-etal-2017-improving}; among others),
document coherence (\citealt{bawden-etal-2018-evaluating, beyer-etal-2021-incoherence}; among others),
and grammatical acceptability in general~\cite{warstadt-etal-2020-blimp, xiang-etal-2021-climp}.

Contrastive evaluation allows for a targeted, automated evaluation of generative models but is restricted to a specific behavioral interface, namely the ranking of pre-defined \textit{minimal pairs}.
However, most models in application areas such as translation or conversation are deployed to produce 1-best sequences, exposing a different behavioral interface to users.
While this limitation of contrastive evaluation is well known, its practical relevance has been unclear.

We show that under certain conditions, the gap between evaluation and deployment can indeed cause misleading results.
As a main factor we identify the \textit{distributional discrepancy} of contrastive evaluation datasets:
Minimal pairs are usually derived from human-written references, but when deployed, a model is conditioned on its own output.

To measure the effect of this factor on evaluation, we focus on neural machine translation (NMT) systems.
Our approach is to test \textit{implausible research hypotheses} in addition to plausible ones.
We find that distributional discrepancy increases the number of false positives regarding implausible hypotheses.
They particularly occur when evaluating distilled NMT models~\cite{kim-rush-2016-sequence}, indicating that in such models, ranking behavior on noisy sequences diverges from generative behavior.

We also propose a way to reduce the distributional discrepancy of minimal pairs.
Our experiments show that false positives can be largely avoided by using machine-generated text instead of human-written text.
This inspires us to release \textbf{\distillingeval}, a variant of the \lingeval English--German MT evaluation suite~\cite{sennrich-2017-grammatical} that uses MT-generated references.

We recommend that future efforts to create contrastive datasets for the evaluation of language generation models minimize distributional discrepancy between evaluation and deployment.
Due to the possibility of false positives, linguistic conclusions about \textit{knowledge} or \textit{abilities} of models should be corroborated by additional evidence from a more natural setting.

\section{Background and Related Work}

\subsection{Contrastive Evaluation}
Contrastive evaluation compares the probability scores that a model assigns to two minimally different sequences.
For example, the sentences \textit{``The cats sleep''} and \textit{``The cats sleeps''} differ in verb number only; if a model assigns higher scores to sentences of the first kind than to sentences of the second kind, it is said to prefer verb forms in agreement with the noun~\cite{linzen-etal-2016-assessing}.

An established method for scoring is to compute the score for a full sentence $X=x_0, x_1, ..., x_n$ as the sum of token log-probabilities predicted by the model $\theta$ ~\cite{marvin-linzen-2018-targeted}:

\[
    \textrm{score}(X)=\sum_{i=0}^{n}\log p_{\theta}(x_{i}|x_{<i})
    \tag{1} \label{eq:score}
\]

When contrastive evaluation is applied to sequence-to-sequence models, two target sequences are scored given the same source sequence~$X$~\cite{sennrich-2017-grammatical}.
We follow previous work and normalize sequence-to-sequence scores by the length of the target sequence $Y$:

\[
    \textrm{score}(Y|X)=\frac{1}{|Y|}\sum_{i=0}^{|Y|}\log p_{\theta}(y_{i}|X,y_{<i})
    \tag{2} \label{eq:seq2seq-score}
\]

\subsection{Limitations of Forced Choice}
Since a limited set of variants is scored, contrastive evaluation presents the model with a forced choice.
In fact, scoring a pre-defined sequence is related to \textit{teacher forcing}, i.e., the conditioning of a model on a ground truth prefix during training.
Whenever an application involves unconstrained generation, a discrepancy between evaluation and deployment arises that is comparable to the \textit{exposure bias} of cross-entropy training~\cite{ranzato2016sequence}.

With regard to syntactic evaluation of language models, \citet{newman-etal-2021-refining} point out that contrastive evaluation and evaluation of systematicity across a paradigm do not necessarily describe a model's likely behavior.
They propose to analyze the complete search space, which, however, is difficult to implement in many use cases.
We pursue a different strategy and create minimal pairs that are more similar to sequences the model will likely generate at deployment time.

\section{Experiments}

In previous work, contrastive evaluation has commonly been used to test \textit{plausible} research hypotheses, for example the hypothesis that RNNs can predict long-distance number agreement~\cite{gulordava-etal-2018-colorless}, or the hypothesis that word dropout improves pronoun resolution in translation~\cite{fernandes-etal-2021-measuring}.
In this paper, we are interested in \textit{implausible} hypotheses and in how the testing of such hypotheses is affected by the limitations described in the previous section.

We formulate two implausible hypotheses about NMT systems, which we mark with an asterisk (*):
\begin{enumerate}
  \item \textit{*Vague language:}\\NMT systems make liberal use of vague placeholder words. Specifically, English–German models use the German placeholder noun \textit{Ding} (`thing') ubiquitously.
  \item \textit{*Hypercorrection:}\\NMT systems have a tendency for hypercorrect language. Specifically, English–German models tend to use genitive case with prepositions that require dative case.
\end{enumerate}

Examples are given in the next section.
The two hypotheses are chosen because they seem implausible both theoretically and empirically.
From a theoretical standpoint, both linguistic phenomena rarely occur in the training data and the model is unlikely to adopt them broadly.
Furthermore, the cognitive and social factors that cause the phenomena in human speech do not apply to neural language models.
Empirically, we find that both phenomena are indeed very rare in neural machine translations, independent of model quality.

For comparison, we also test two plausible hypotheses about NMT systems:

\begin{enumerate} \setcounter{enumi}{2}
  \item \textit{Polarity affix deletion:}\\NMT systems sometimes omit negation affixes, changing the polarity of a word~\cite{hossain-etal-2020-non}. Specifically, English–German models sometimes omit the negation prefix \textit{un-} from German words~\cite{sennrich-2017-grammatical}.
  \item \textit{Clause omission:}\\NMT systems sometimes omit a clause from the translated sentence~\cite{tu-etal-2016-modeling}.
\end{enumerate}

\subsection{Test Set Creation}\label{subsec:test-set-creation}

For each of the four hypotheses, we create an English–German contrastive test set.
For \textit{vague language}, \textit{polarity affix deletion} and \textit{clause omission}, we use the newstest datasets 2009--2016 as a data source. For \textit{hypercorrection}, we combine five data sources: newstest 2009--2019 as well as OpenSubtitles2016~\cite{lison-tiedemann-2016-opensubtitles2016}, TED2020~\cite{reimers-gurevych-2020-making}, QED~\cite{abdelali-etal-2014-amara} and JW300~\cite{agic-vulic-2019-jw300}, which are provided by OPUS~\cite{tiedemann-2012-parallel}.

\paragraph{Vague language}
Contrastive variants are created by replacing a random noun in each reference with an uninflected \textit{Ding} `thing', which is a common replacement noun in spoken German~\cite{vogel_2020}:

\begin{itemize}[noitemsep, leftmargin=0pt]
  \item[] English: \textit{Prague Stock Market falls to minus by the \textbf{end of the trading day}}
  \item[] German (correct): \textit{Die Prager Börse stürzt gegen \textbf{Geschäftsschluss} ins Minus.}
  \item[] German (contrastive): \textit{Die Prager Börse stürzt gegen \textbf{Ding} ins Minus.}
\end{itemize}

\paragraph{Hypercorrection}
To create contrastive variants for hypercorrect genitives, we select references containing German propositions that require dative in Standard German, but are sometimes used hypercorrectly with a genitive case~\cite{HentschelWeydt+2013}.\footnote{We use the prepositions \textit{entgegen}, \textit{entsprechend}, \textit{gegen\-über}, \textit{gemäß}, \textit{nahe}, \textit{nebst}, \textit{(mit)samt} and \textit{seit}.}
We construct contrastive variants by converting the dative case into genitive case:

\begin{itemize}[noitemsep, leftmargin=0pt]
  \item[] English: \textit{I've loved you ever \textbf{since that day} in the rose garden.}
  \item[] German (correct): \textit{Ich liebe dich \textbf{seit dem Tag} im Rosengarten.}
  \item[] German (contrastive): \textit{Ich liebe dich \textbf{seit des Tags} im Rosengarten.}
\end{itemize}

\paragraph{Polarity affix deletion}
Contrastive variants are created by deleting the prefix \textit{un-} from adjectives, adverbs and nouns in the German references in cases where this changes the polarity of the word, similar to the test set created by \citet{sennrich-2017-grammatical}:

\begin{itemize}[noitemsep, leftmargin=0pt]
  \item[] English: \textit{The probes \textbf{unexpectedly} become faster or slower.}
  \item[] German (correct): \textit{Die Sonden werden \textbf{unerwartet} schneller oder langsamer.}
  \item[] German (contrastive): \textit{Die Sonden werden \textbf{erwartet} schneller oder langsamer.}
\end{itemize}

\paragraph{Clause omission}
Contrastive variants are created by deleting a clause from the reference.
As clauses we treat token sequences segmented by the Stanza sentence splitter~\cite{qi-etal-2020-stanza}:

\begin{itemize}[noitemsep, leftmargin=0pt]
  \item[] English: \textit{And even if it could be proved for humans - \textbf{how would one want to prove it for rats?}}
  \item[] German (correct): \textit{Und selbst wenn man das für den Menschen beweisen könnte: \textbf{Wie wollte man es bei Ratten nachweisen?}}
  \item[] German (contrastive): \textit{Und selbst wenn man das für den Menschen beweisen könnte:}
\end{itemize}

\subsection{Human-Written References}
The above test sets are derived from naturally occurring parallel text, which is common practice when creating contrastive datasets for MT~\cite{sennrich-2017-grammatical, rios-gonzales-etal-2017-improving, bawden-etal-2018-evaluating, muller-etal-2018-large, voita-etal-2019-good, raganato-etal-2019-mucow, sugiyama-yoshinaga-2019-data, nagata-morishita-2020-test, shimazu-etal-2020-evaluation, lopes-etal-2020-document, he-etal-2020-box, stojanovski-etal-2020-contracat}.
However, comparisons have shown that human-written references are different from machine translations in that they contain more noise and have more linguistic diversity~\cite{zhang2018analyzing, vanmassenhove-etal-2019-lost}.

We propose to measure the ``distance'' between a pre-defined target sequence and the 1-best translation~$\hat{Y}$ generated by an MT system as the difference in log-scores (according to Equation \ref{eq:seq2seq-score}) that the system assigns to the two sequences.
Furthermore, we define the \textit{distributional discrepancy} of a contrastive evaluation dataset as the mean difference in scores between the 1-best translation and the preferred variant:

\small
\begin{equation*}
\begin{aligned}
\textrm{score\_preferred} ={} & \max(\textrm{score}(Y^{correct}),\textrm{score}(Y^{contrast.})) \\
\textrm{discrepancy} ={} & \frac{1}{n}\sum_{i=0}^{n}\textrm{score}(\hat{Y}_{i})-\textrm{score\_preferred}_{i}
\end{aligned}
\end{equation*}
\normalsize
It should be noted that this definition of distributional discrepancy is mainly useful for comparing multiple test sets with respect to a single model.
It is less useful for assessing a single test set with respect to multiple models, because score differences are not necessarily comparable between models.

\begin{table*}[]
\begin{tabularx}{\textwidth}{@{}XXrrrr@{}}
 \textbf{Hypothesis}                        & \textbf{Test set type}                        & \multicolumn{2}{r}{\textbf{Discrepancy of test set}}      & \multicolumn{2}{r@{}}{\textbf{Reported accuracy}}                              \\
                              &  & \multicolumn{1}{r}{\textsc{Transformer}} & \multicolumn{1}{r}{\textsc{Distilled}} & \multicolumn{1}{r}{\textsc{Transformer}} & \multicolumn{1}{r@{}}{\textsc{Distilled}} \\ \midrule
\textit{*Vague language} & \makebox[0pt][l]{human references}      & $1.2\pm0.0$  & $2.5\pm0.1$   & $99.1\pm0.1$  & $94.7\pm0.4$     \\
 & \makebox[0pt][l]{machine references}                            & $0.3\pm0.0$  & $0.7\pm0.0$   & $99.9\pm0.0$  & $98.7\pm0.2$  \\ \midrule
\textit{*Hypercorrection} & \makebox[0pt][l]{human references}     & $1.3\pm0.0$  & $2.7\pm0.1$   & $95.4\pm0.3$  & $91.2\pm0.5$     \\
 & \makebox[0pt][l]{machine references}                            & $0.4\pm0.0$  & $1.1\pm0.1$   & $99.9\pm0.1$  & $99.6\pm0.4$  \\ \midrule
\textit{Polarity affix del.} & \makebox[0pt][l]{human references}  & $1.3\pm0.0$  & $2.7\pm0.1$   & $94.0\pm1.1$  & $78.3\pm0.9$     \\
& \makebox[0pt][l]{machine references}                             & $0.3\pm0.0$  & $0.7\pm0.1$   & $96.7\pm1.5$  & $93.9\pm1.1$  \\ \midrule
\textit{Clause omission} & \makebox[0pt][l]{human references}      & $1.3\pm0.0$  & $2.8\pm0.1$   & $75.5\pm3.7$  & $71.3\pm0.7$     \\
& \makebox[0pt][l]{machine references}                             & $0.3\pm0.0$  & $0.7\pm0.0$   & $87.7\pm2.7$  & $86.3\pm2.7$  \\ \bottomrule
\end{tabularx}
\caption{Results for four different hypotheses about English--German NMT systems. An asterisk (*) marks hypotheses that are a priori implausible. The table reports distributional discrepancies of different test set types, as well as the accuracy scores achieved by non-distilled and distilled systems when evaluated with the test sets. We report averages and standard deviations across three models trained independently with different random seeds.}
\label{tab:results}
\end{table*}

\subsection{Machine-Generated References}\label{subsec:machine-generated-references}
With the goal of reducing distributional discrepancy, we create versions of our test sets that use machine-generated references.
First, we re-translate the sources from our test sets using commercial NMT systems.\footnote{We used \textit{Amazon Translate}, \textit{DeepL Translator}, \textit{Google Translate}, and \textit{Microsoft Translator} for 25\% sentences each.}
We then repeat the steps described in Section~\ref{subsec:test-set-creation} to create contrastive variants.

\paragraph{Validation}
Since some machine-generated references contain errors, a validation step is needed.
The validation should ensure that (a) the machine references are correct with respect to the linguistic phenomenon at hand, and that (b) no undesired bias is introduced into the evaluation.

We use a semi-automatic approach and look for lexical overlap with the human references regarding the phenomenon.
For example, in the case of polarity affix deletion, we label the machine reference as correct if it contains the same polarity word as the human reference.
Otherwise we manually check whether the machine reference might be incorrect, but only if it contains the same polarity word as the human contrastive variant.
This occurs rarely, and most of the time we find that it is the original human reference that is incorrect while the machine reference is correct.
In the rare cases where the machine reference is verifiably incorrect with regard to the phenomenon, we use it as the contrastive variant and derive the correct variant manually.

Machine references that have no phenomenon-specific lexical overlap to the human references are dropped from the test set because they cannot be automatically validated.
This raises the question whether test sets created in such a way contain undesired bias.

\paragraph{Dataset Bias}
We discuss two kinds of bias that might be introduced.
First of all, by only including machine references that can be classified automatically as either correct or incorrect based on the human references, the distribution of the machine-generated test set could become more similar to the human-written test set.
However, our experiments show that the difference in distributional discrepancy between the two test sets is sufficiently large.
Future work could avoid this bias by employing human annotators to validate machine references.

Secondly, it might be that machine references only use the phenomenon in unambiguous contexts.
This would cut off the long tail of human-written test samples that is especially challenging for NLP models.
While such a bias is likely to be introduced to a degree, we see it is a \textit{desired} bias, since our goal is to reduce distributional discrepancy between a test set and the generative behavior of an evaluated system.

\subsection{Experimental Setting}
We evaluate two types of NMT systems:
\begin{enumerate}[noitemsep]
  \item \textsc{Transformer}: Transformer models of size `big'~\cite{NIPS2017_3f5ee243}.
  \item \textsc{Distilled}: Transformer models of size `small' distilled from (1) using sequence-level knowledge distillation~\cite{kim-rush-2016-sequence}.
\end{enumerate}

\paragraph{Training}
For both types, we trained three models with different random seeds.
To train the \textsc{Transformer} models, we used similar data and configuration as~\citet{ng-etal-2019-facebook}, using Fairseq~\cite{ott-etal-2019-fairseq}.
We used the English--German parallel training data from the WMT19 news translation task~\cite{barrault-etal-2019-findings}.
Sentences longer than 250 tokens and pairs with a length ratio larger than 1.5 were filtered, resulting in 42.9M sentence pairs used for training and distillation.
We selected the best checkpoint with respect to BLEU based on the \textit{newstest} sets from the preceding years.

\paragraph{Distillation}
We then used each of the three \textsc{Transformer} models as a teacher to train an individual student model.
A comparison of hyperparameters is provided in Appendix~\ref{sec:hyperparameters}.

For decoding we always use beam search with size 5.

\paragraph{Model Quality}
The models of type \textsc{Transformer} achieve an average BLEU score of $37.3\pm0.3$, while the \textsc{Distilled} models achieve $35.7\pm0.4$ BLEU when evaluated on newstest19.

\subsection{Results}
The left-hand side of Table~\ref{tab:results} shows the distributional discrepancies of the test sets.
As expected, the test sets derived from human-written references have a higher discrepancy, while those derived from machine-generated references are closer to what the evaluated model would generate.

The right-hand side of Table~\ref{tab:results} shows the reported \textit{accuracy} of the evaluated models, i.e.\ the ratio of test instances where the model prefers the correct variant over the contrastive variant.
While all the accuracies are much better than random, the results for implausible hypotheses seem to indicate that models do occasionally generate the implausible phenomena, and that distilled models generate them more often than other models.
Since this is not reflected by the actual generative behavior, the testing of implausible hypotheses shows the danger of false positives.

The test sets with machine-generated references produce far fewer false positives.
The reported accuracy is higher with machine references also for the plausible hypotheses, but a gap to 100\% accuracy remains, which is in line with previous work on these types of NMT errors~\cite{hossain-etal-2020-non, tang2021revisiting, tu-etal-2016-modeling}.

\section{Dataset Release}
Given the improved specificity of test sets with machine-generated references, we release corresponding test sets for other phenomena in the \lingeval test suite~\cite{sennrich-2017-grammatical}, terming our dataset variant \textbf{\distillingeval}.

\lingeval, the original test suite, is a collection of 97k contrastive translation pairs for 13 different error types in English--German translation.
Building on \lingeval, we create test sets with machine-generated references for the following error types, in addition to the ones discussed in the previous section: \textit{noun phrase agreement}, \textit{subject-verb agreement} and other \textit{polarity deletion} phenomena involving the German negation lexemes \textit{kein} and \textit{nicht}.
Results for these test sets are reported in Table~\ref{tab:distillingeval-results}, and further results for a state-of-the-art NMT system are provided in Appendix~\ref{sec:baseline-results}.
Table~\ref{tab:dataset-sizes} provides an overview of the test set sizes per error type in \distillingeval.

\begin{table}[]
\begin{tabularx}{\columnwidth}{@{}Xrr@{}}
Error type                  & Human Ref. & MT Ref. \\ \midrule
\textit{clause\_omission}             & 1104               & 1025                 \\
\textit{hypercorrect\_genitive}       & 3404               & 635                  \\
\textit{np\_agreement}                & 24\,055              & 10\,595                \\
\textit{placeholder\_ding}            & 18\,647              & 18\,659                \\
\textit{polarity\_affix\_del}          & 408                & 180                  \\
\textit{polarity\_particle\_kein\_del}  & 554                & 201                  \\
\textit{polarity\_particle\_nicht\_del} & 2561               & 888                  \\
\textit{subj\_verb\_agreement}         & 31\,978              & 6701
            \\ \bottomrule
\end{tabularx}
\caption{Number of samples per \distillingeval error type. Error types with machine-generated references tend to have fewer samples, which is discussed in Section~\ref{subsec:machine-generated-references}.}
\label{tab:dataset-sizes}
\end{table}

\begin{table*}[]
\begin{tabularx}{\textwidth}{@{}X@{\hskip 40pt}Xrrrr@{}}
 \textbf{Error Type}                        & \textbf{Test set type}                        & \multicolumn{2}{r}{\textbf{Discrepancy of test set}}      & \multicolumn{2}{r@{}}{\textbf{Reported accuracy}}                              \\
                              &  & \multicolumn{1}{r}{\textsc{Transf.}}  & \multicolumn{1}{r}{\textsc{Distilled}} & \multicolumn{1}{r}{\textsc{Transf.}} & \multicolumn{1}{r@{}}{\textsc{Distilled}} \\ \midrule
\textit{np\_agreement} & \makebox[0pt][l]{human references}                  & $2.0\pm0.6$  & $2.6\pm0.1$   & $95.9\pm2.5$  & $94.4\pm0.8$     \\
 & \makebox[0pt][l]{machine references}                                      & $0.5\pm0.2$  & $0.7\pm0.1$   & $98.8\pm0.8$  & $99.0\pm0.4$  \\ \midrule
\textit{polarity\_particle\_kein\_del} & \makebox[0pt][l]{human references}  & $1.3\pm0.0$  & $2.7\pm0.1$   & $95.3\pm0.8$  & $90.7\pm0.9$     \\
 & \makebox[0pt][l]{machine references}                                      & $0.2\pm0.0$  & $0.6\pm0.1$   & $99.8\pm0.3$  & $99.8\pm0.3$  \\ \midrule
\textit{polarity\_particle\_nicht\_del} & \makebox[0pt][l]{human references} & $1.3\pm0.0$  & $2.6\pm0.1$   & $95.5\pm0.2$  & $87.9\pm0.7$     \\
& \makebox[0pt][l]{machine references}                                       & $0.3\pm0.0$  & $0.7\pm0.0$   & $99.7\pm0.3$  & $98.8\pm0.1$  \\ \midrule
\textit{subj\_verb\_agreement} & \makebox[0pt][l]{human references}          & $1.2\pm0.0$  & $2.6\pm0.1$   & $97.1\pm0.3$  & $91.4\pm0.3$     \\
& \makebox[0pt][l]{machine references}                                       & $0.3\pm0.0$  & $0.7\pm0.0$   & $99.2\pm0.1$  & $97.9\pm0.3$  \\ \bottomrule
\end{tabularx}
\caption{Test set discrepancies and model accuracies for the other four error types included in \distillingeval (in addition to the four error types in Table~\ref{tab:results}).}
\label{tab:distillingeval-results}
\end{table*}

\section{Discussion}
By testing implausible hypotheses, we demonstrate the risk of drawing wrong inferences about generative behavior of (conditional) language models, especially if there is a large distributional discrepancy between minimal pairs and generated sequences.

This problem is especially apparent for distilled NMT models, which perform poorly on human-written minimal pairs because they were never exposed to such a distribution during training.
While this indicates that distilled NMT models are less robust against improbable contexts, human-crafted minimal pairs also become less useful to predict their unconstrained generative behavior.

The danger of false positives from minimal pairs highlights the fact that
behaviorist approaches to measuring knowledge are limited to the behavioral interface that is observed.
Systematic assessments of linguistic \textit{knowledge} or syntactic \textit{abilities} of neural models should be qualified accordingly, in case minimal pairs are the primary analysis method.
We suspect that whenever a broad range of hypotheses is tested, including phenomena that are rarely observed in actual machine-generated text, the risk of false positives is increased.

We thus recommend that minimal pairs be constructed from machine-generated text in evaluation settings where unconstrained generation is the behavior of interest.
This is relatively straightforward for sequence-to-sequence evaluation, as we demonstrated in our experiments.
For other settings, e.g.~the evaluation of dialogue models, obtaining useful machine-generated text might require more elaborate techniques, such as round-trip translation.

However, human-crafted minimal pairs remain valuable in other use cases.
While machine-generated pairs may be more appropriate when the main interest is to study the model's behavior close to its mode, e.g.\ in a sequence-to-sequence task, human-written pairs (or pairs that are machine-generated to be different from the training distribution on purpose) may tell us more about the robustness of models outside the mode.
For example, terminology-constrained or interactive applications depend on robustness against improbable contexts, and contrastive evaluation indicates that current NMT systems lack such robustness~\cite{stojanovski-etal-2020-contracat}.
Similarly, syntactic evaluation of language models using randomly generated or nonsensical sentences~\cite{gulordava-etal-2018-colorless, warstadt-etal-2020-blimp} can be seen as method to assess the robustness of a model under improbable input, rather than as an assessment of generative capabilities in general.

\section{Conclusion}
We show that there are conditions where contrastive evaluation leads to false positives if generative behavior is inferred from behavior under forced choice.
Experiments with English--German NMT indicate that the gap between the two behavioral interfaces is especially high when human-written text is used to create minimal pairs.
Using machine-generated text largely reduces the gap.
We recommend that human-written minimal pairs are mainly used for assessing the robustness of models, but that for predicting the generative behavior of language models, machine-generated minimal pairs are used.

\subsection*{Broader Impact}

For language generation systems to be deployed, they should behave according to specified principles in a robust way.
Typical requirements are linguistic acceptability, avoidance of undesirable societal biases~\cite{sheng-etal-2021-societal}, and the avoidance of harmful speech acts.
Contrastive evaluation is one of several methods that can help predict the behavior of language generation systems.
However, to our knowledge the method has been mainly used to evaluate linguistic acceptability, and less to evaluate ethically sensitive aspects of generation.

It is crucial that evaluation methods have a high predictiveness regarding the behavior of a deployed system.
On the one hand, lack of sensitivity can lead to unforeseen negative impact.
On the other hand, lack of specificity -- which we address in this paper -- reduces the usefulness of comparisons between systems.

\section*{Acknowledgments}
This work was funded by the Swiss National Science Foundation (project MUTAMUR; no.~176727) and made use of infrastructure services provided by S3IT, the Service and Support for Science IT team at the University of Zurich.
We would like to thank Anne Beyer and the anonymous reviewers for helpful feedback.

\bibliography{bibliography}
\bibliographystyle{acl_natbib}

\appendix

\clearpage
\onecolumn

\section{Hyperparameters}\label{sec:hyperparameters}

\begin{table*}[htb!]
\begin{tabularx}{\textwidth}{@{}Xrrrrr@{}}
\toprule
Name  & $N$ & $d_{\text{model}}$ & $d_{\text{ffn}}$  & $h$  & Parameters  \\ \midrule
\textsc{Transformer} (big)   & 6 & 1024   & 8192 & 16 & 269.7M \\
\textsc{Distilled} (small) & 4 & 512    & 2048 & 4  & 50.9M \\ \bottomrule
\end{tabularx}
\caption{Hyperparameters of the Transformer variants used for the experiments}
\label{tab:transformer-sizes}
\end{table*}

\section{Examples}\label{sec:examples}

\begin{table*}[htb!]
\begin{tabularx}{\textwidth}{@{}X@{\hskip 20pt}r@{}}
\toprule
\textbf{Example Inputs (English--German)} & \makebox[0pt][r]{\textbf{Score Assigned by Model}} \\

\midrule
Source: \textit{Yesterday evening, the committee wanted to vote on the \textbf{appointment}.} & \\
[4pt] 1-best translation by the evaluated system: \\\textit{Gestern Abend wollte der Ausschuss über die \textbf{Ernennung} abstimmen.} & -0.09 \\
[4pt] Minimal pair based on a human-written reference: & \\
[2pt] -- Correct: \textit{Gestern Abend wollte das Gremium über die \textbf{Personalie} abstimmen.} & -3.61 \\
[2pt] -- Incorrect: \textit{Gestern Abend wollte das Gremium über die \textbf{Ding} abstimmen.} & -2.34 \\
[4pt] Minimal pair based on a machine-generated reference (Amazon Translate): & \\
[2pt] -- Correct: \textit{Gestern Abend wollte der Ausschuss über die \textbf{Ernennung} abstimmen.} & -0.09 \\
[2pt] -- Incorrect: \textit{Gestern Abend wollte der Ausschuss über die \textbf{Ding} abstimmen.} & -1.25 \\

\midrule
Source: \textit{Why did Judah lose its land \textbf{and} temple?} & \\
[4pt] 1-best translation by the evaluated system: &  \\
\textit{Warum hat Juda sein Land \textbf{und} seinen Tempel verloren?} & -0.11 \\
[4pt] Minimal pair based on a human-written reference: & \\
[2pt] -- Correct: \textit{Warum verlor Juda sein Land \textbf{mitsamt dem} Tempel?} & -2.58 \\
[2pt] -- Incorrect: \textit{Warum verlor Juda sein Land \textbf{mitsamt des} Tempels?} & -2.55 \\
[4pt] Minimal pair based on a machine-generated reference (DeepL): & \\
[2pt] -- Correct: \textit{Warum hat Juda sein Land \textbf{und} seinen Tempel verloren?} & -0.11 \\
[2pt] -- Incorrect: N/A & \\

\bottomrule
\end{tabularx}
  \caption{Examples of human-written and machine-generated minimal pairs for the \textit{*Vague language} hypothesis (top) and the \textit{*Hypercorrection} hypothesis (bottom).
  The log-scores are computed by an NMT model of type \textsc{Distilled}.\\
      The first example demonstrates that a model often assigns a lower score to the correct human reference than to the incorrect machine reference.
  The human reference differs from the machine reference only in how the words \textit{committee} and \textit{appointment} are translated.
 The human word choice is fluent but has a lower probability under the model.
  \\
      The second example shows that machine references often avoid the phenomenon altogether.
  Here, a simple conjunction is used instead of the more prestigious preposition \textit{mitsamt} `along with' in the human reference.
  This removes any risk of inserting a hypercorrect genitive.
  Since a contrastive variant cannot be derived from the machine reference, the sample is excluded from the machine-generated test set.
}
\label{tab:examples-hypercorrection}
\end{table*}

\clearpage
\section{State-of-the-art Accuracies for \distillingeval}\label{sec:baseline-results}

\begin{table*}[htb!]
\begin{tabularx}{\textwidth}{@{}X@{\hskip 40pt}Xrrrr@{}}
\toprule
 \textbf{Error Type}                        & \textbf{Test set type}                        & \textbf{Discrepancy of test set}      & \textbf{Reported accuracy} \\ \midrule
\textit{clause\_omission} & \makebox[0pt][l]{human references}               & $0.91$    & $78.1$      \\
 & \makebox[0pt][l]{machine references}                                      & $0.19$    & $87.1$   \\ \midrule
\textit{hypercorrect\_genitive} & \makebox[0pt][l]{human references}         & $1.13$    & $94.2$      \\
 & \makebox[0pt][l]{machine references}                                      & $0.18$    & $100.0$   \\ \midrule
\textit{np\_agreement} & \makebox[0pt][l]{human references}                  & $0.84$    & $99.7$      \\
 & \makebox[0pt][l]{machine references}                                      & $0.15$    & $100.0$   \\ \midrule
\textit{placeholder\_ding} & \makebox[0pt][l]{human references}              & $0.79$    & $99.8$      \\
 & \makebox[0pt][l]{machine references}                                      & $0.16$    & $100.0$   \\ \midrule
\textit{polarity\_affix\_del} & \makebox[0pt][l]{human references}           & $0.87$    & $98.8$      \\
 & \makebox[0pt][l]{machine references}                                      & $0.13$    & $100.0$   \\ \midrule
\textit{polarity\_particle\_kein\_del} & \makebox[0pt][l]{human references}  & $0.88$    & $97.5$      \\
 & \makebox[0pt][l]{machine references}                                      & $0.12$    & $100.0$   \\ \midrule
\textit{polarity\_particle\_nicht\_del} & \makebox[0pt][l]{human references} & $0.84$    & $97.9$      \\
& \makebox[0pt][l]{machine references}                                       & $0.14$    & $99.9$   \\ \midrule
\textit{subj\_verb\_agreement} & \makebox[0pt][l]{human references}          & $0.83$    & $98.7$      \\
& \makebox[0pt][l]{machine references}                                       & $0.14$    & $99.8$   \\ \bottomrule
\end{tabularx}
\caption{\distillingeval results of a state-of-the-art NMT ensemble~\cite{ng-etal-2019-facebook}.
The accuracies on machine references suggest that clause omission is an error type that still occurs with state-of-the-art NMT systems.
}
\label{tab:sota-results}
\end{table*}

\end{document}